\documentclass[twoside]{article}

\usepackage{natbib} 
\usepackage{amsmath}
\usepackage{amssymb}
\usepackage{graphicx}
\usepackage{subfig}
\usepackage{placeins}

\usepackage[utf8]{inputenc} 
\usepackage[T1]{fontenc}    
\usepackage{hyperref}       
\usepackage{url}            
\usepackage{booktabs}       
\usepackage{amsfonts}       
\usepackage{nicefrac}       
\usepackage{microtype}      
\usepackage{xcolor}         
\usepackage{graphicx}
\usepackage{amsmath}
\usepackage{amssymb}
\usepackage{mathtools}
\usepackage{amsthm}
\usepackage{microtype}
\usepackage{graphicx}
\usepackage{booktabs} 
\usepackage{wrapfig}
\usepackage[ruled,vlined]{algorithm2e}

\theoremstyle{plain}
\newtheorem{theorem}{Theorem}[section]

\theoremstyle{definition}

\theoremstyle{remark}
\newtheorem{remark}[theorem]{Remark}

\makeatletter
\def\Notice@String{}
\makeatother

\usepackage[accepted]{aistats2025}
\begin{document}

%

%

\twocolumn[

\aistatstitle{Consistent Labeling Across Group Assignments: 
\\Variance Reduction in Conditional Average Treatment Effect Estimation}

\aistatsauthor{ Yi-Fu Fu \And Keng-Te Liao \And  Shou-De Lin }

\aistatsaddress{ National Taiwan University \And  National Taiwan University \And National Taiwan University } ]

\begin{abstract}
Numerous algorithms have been developed for Conditional Average Treatment Effect (CATE) estimation. In this paper, we first highlight a common issue where many algorithms exhibit inconsistent learning behavior for the same instance across different group assignments. We introduce a metric to quantify and visualize this inconsistency. Next, we present a theoretical analysis showing that this inconsistency indeed contributes to higher test errors and cannot be resolved through conventional machine learning techniques. To address this problem, we propose a general method called \textbf{Consistent Labeling Across Group Assignments} (CLAGA), which eliminates the inconsistency and is applicable to any existing CATE estimation algorithm. Experiments on both synthetic and real-world datasets demonstrate significant performance improvements with CLAGA.
\end{abstract}

\section{INTRODUCTION}
\label{sec:intro}
In recent years, personalized decision-making has become increasingly prevalent, spanning areas such as medicine \citep{bica2021real, foster2011subgroup, jaskowski2012uplift},  marketing \citep{gubela2019conversion}, and sociology \citep{sociology_ref}. A critical component of personalized decision-making is understanding how different individuals or subgroups respond differently to interventions. This is where the Conditional Average Treatment Effect (CATE) plays a central role. CATE estimates the expected effect of a treatment conditional on certain characteristics of the individual or subgroup, allowing for targeted and effective interventions.

Estimating CATE accurately from real-world data, however, is a challenging task. Existing CATE estimation methods typically follow the potential outcomes framework \citep{rubin1974estimating, rubin2005causal}, which quantifies treatment effects by comparing the outcomes an individual would have experienced if they had received the treatment versus if they had not. The main challenge in estimating CATE lies in the fact that we can only observe the outcome under one condition—either with or without the treatment—making it impossible to directly measure the treatment effect for an individual.

To address this challenge, machine learning approaches have adopted various strategies \citep{zhang2021unified, gutierrez2017causal}, with the goal of generating learning targets that can (a) facilitate supervised learning for training estimators, and (b) ensure unbiased estimation. Among the early methods, both the single-model approach \citep{lo2002true} and the two-model approach \citep{hansotia2002incremental} have remained competitive over the years. More recently, advanced techniques such as the X-learner \citep{kunzel2019metalearners}, R-learner \citep{nie2021quasi}, and Doubly-Robust learner \citep{kennedy2020optimal} have been proposed. These modern methods also follow the principle of transforming the CATE estimation problem into one or more supervised learning tasks.

Different from prior works, which primarily focus on unbiased estimation, this paper emphasizes the variance and statistical efficiency of CATE estimators. Our contributions are threefold:

\begin{enumerate} \item \textbf{Identification of Inconsistent Learning across Group Assignments}: We reveal the problem of inconsistent learning behavior in CATE estimation algorithms, where the prediction for a given training instance differs significantly depending on whether it is assigned to the treatment group or the control group. This inconsistency persists even with large datasets, up to 1 million instances. To quantify this effect, we introduce a metric, termed the \emph{discrepancy ratio}.

\item \textbf{Theoretical Framework for Error Decomposition}: We propose a theoretical framework that decomposes CATE estimation errors into distinct components, with a subset identified as being linked to group assignment inconsistencies. We show that these inconsistencies indeed contribute to higher estimation errors and are inherent to algorithm design, making them resistant to traditional machine learning solutions such as hyperparameter tuning or model selection.

\item \textbf{Proposed Solution for Consistent Labeling}: We introduce a method called \emph{consistent labeling across group assignments} (CLAGA), which is applicable to any off-the-shelf CATE algorithm. By applying this method, the error term related to inconsistent learning across group assignments is effectively eliminated. Our experimental results, conducted on both synthetic and real-world datasets, demonstrate significant performance improvements, supporting the validity and practical benefits of our proposed method.
\end{enumerate}

\section{PROBLEM FORMULATION}
\label{sec: problem formulation}
In this paper, we follow the potential outcomes framework proposed by \cite{rubin1974estimating, rubin2005causal}. In this framework, each instance is represented by a feature vector \(X \in \mathcal{X}\). Each instance \(X_i\) is assigned a treatment indicator \(W_i \in \{0,1\}\), which indicates whether the treatment is received. Additionally, each instance has two potential outcomes, \(Y_i^{(0)}\) and \(Y_i^{(1)}\), which represent the outcome if the treatment is not received (\(W_i=0\)) and if the treatment is received (\(W_i=1\)), respectively.

We adopt the standard assumptions of ``consistency'', ``unconfoundedness'', and ``positivity'' within the potential outcomes framework \citep{rubin2005causal}, which are defined as follows:

\begin{itemize}
  \item \textbf{Consistency}: The observed outcome \(Y\) is equal to the potential outcome corresponding to the received treatment assignment:
  \[
    Y = 
    \begin{cases}
      Y^{(1)} & \text{if } W = 1, \\
      Y^{(0)} & \text{if } W = 0.
    \end{cases}
  \]
    
    \item \textbf{Unconfoundedness}: Conditional on covariates $X$, the treatment assignment is independent of the potential outcomes:
    \begin{equation*}
    (Y^{(0)}, Y^{(1)}) \perp\!\!\!\perp W \mid X
    \end{equation*}    
    
  \item \textbf{Positivity (Overlap)}: Each unit has a non-zero probability of receiving either treatment:
  \[
    0 < \Pr(W = 1 \mid X = x) < 1 \quad \text{for all } x \in \mathcal{X}.
  \]
\end{itemize}

The conditional potential outcome functions, \(\mu_0(x)\) and \(\mu_1(x)\), denote the expected outcomes under the two group assignments. Specifically, \(\mu_0(x) = \mathbb{E}[Y^{(0)} \mid X = x]\) and \(\mu_1(x) = \mathbb{E}[Y^{(1)} \mid X = x]\). The Conditional Average Treatment Effect (CATE), \(\tau(x)\), is defined as the difference between these two potential outcome functions:
\begin{equation}\label{eq:definition of tau}
  \tau(x) = \mu_1(x) - \mu_0(x).
\end{equation}


In a typical CATE estimation scenario, we are given a dataset 
\(\mathcal{D} = \{(X_i, W_i, Y_i)\}_{i=1}^n\), where each unit \(i\) is drawn i.i.d. 
from a joint distribution over \((X, W, Y)\), and the observed outcome satisfies 
\(Y_i = Y_i^{(W_i)}\). Since only one of the two potential outcomes 
\((Y_i^{(0)}, Y_i^{(1)})\) can be observed for each individual, direct supervision 
of the treatment effect function \(\tau(x)\) using standard supervised learning 
methods is infeasible.

Note that \(\tau(x) = \mu_1(x) - \mu_0(x)\) is a deterministic function of the covariates \(x\); 
it is not a random variable and does not depend on treatment assignment. 
The central challenge of CATE estimation lies in inferring this function from only 
partial observations of the potential outcomes.

Although we list the standard assumptions of the potential outcomes framework for completeness, our analysis and proposed method do not rely on the unconfoundedness assumption. 
In particular, our method applies to both randomized controlled trials (RCTs), where unconfoundedness holds by design, and observational settings, where the assumption may not be plausible.
We only require that the dataset $(X_i, W_i, Y_i)$ is drawn i.i.d. from a joint distribution, and that the observed outcome satisfies $Y_i = Y_i^{(W_i)}$.

While different algorithms employ a variety of strategies to address this challenge, in Section~\ref{subsec:biased learning}, we demonstrate that most of these strategies lead to inconsistent learning across group assignments. In Section~\ref{Sec:error-term-breakdown}, we further show that such inconsistencies contribute to higher CATE estimation errors.

\section{RELATED WORK}
\label{Sec: related work}

\subsection{Algorithms for CATE Estimation}
\label{Sec: related work: Algorithms for CATE estimation}

CATE estimation is challenging due to the fact that we only observe one of the two potential outcomes for each instance. To address this, various strategies have been developed to enable models to learn from the incomplete information. In this paper, we consider state-of-the-art and widely-used CATE estimation algorithms, including the Single-model approach \citep{lo2002true}, the Two-model approach \citep{hansotia2002incremental,radcliffe2007using}, the X-learner \citep{kunzel2019metalearners}, the R-learner \citep{nie2021quasi}, and the Doubly-Robust (DR) learner \citep{kennedy2020optimal}.

Beyond algorithm design, recent work has proposed improvements to CATE estimation from different angles. For example, \cite{johansson2016learning} introduced regularization to promote similarity in representations between the treatment and control groups. \cite{yao2018representation} proposed preserving local similarity in the feature space. \cite{zhang2021treatment} developed a method to disentangle covariate factors to improve CATE estimation.

\subsection{Comparison of Different Algorithms}
Comparing and selecting among CATE estimation algorithms is an ongoing challenge, as no single method dominates across all data scenarios \citep{dorie2019automated}. Several studies have constructed benchmarks using synthetic and real-world datasets to assess algorithmic performance under diverse conditions \citep{wendling2018comparing, knaus2022double, schuler2017synth}. Other works focus on model selection and performance evaluation criteria, including recent proposals for theoretically grounded metrics \citep{alaa2019validating, curth2021nonparametric}.

While these comparisons emphasize aggregate accuracy or bias metrics, less attention has been paid to whether algorithms learn consistent label functions across treatment groups—a focus of our work. Rather than proposing a new CATE estimator, our approach complements existing algorithms by addressing variance-related inconsistencies that arise even in large samples.


\section{INCONSISTENT LEARNING ACROSS GROUP ASSIGNMENTS}
\label{subsec:biased learning}

In this section, we introduce and analyze an overlooked phenomenon in CATE estimation, which we term \emph{inconsistent learning across group assignments}. Many CATE estimation algorithms have been developed to address the challenge that the ground truth treatment effect \(\tau(x)\) cannot be directly observed. These algorithms often claim to satisfy \emph{pointwise unbiasedness}, meaning that 
\(\mathbb{E}_{M,W}[\hat{\tau}(x)] = \tau(x)\), where the expectation is over 
both training randomness and treatment assignment. 
However, it does not ensure that the learned function is \emph{consistent} when the same individual appears in different treatment groups across runs. In particular, it does not guarantee that the conditional expectations 
\(\mathbb{E}[\hat{\tau}(x) \mid X = x, W = 0]\) and 
\(\mathbb{E}[\hat{\tau}(x) \mid X = x, W = 1]\) are consistent with each other.

Ideally, for a consistent estimator of \(\tau(x)\), the learned value 
should not depend on the group assignment of the instance used during training. 
That is, for any fixed covariate \(x\), the expected estimate should satisfy:
\[
\mathbb{E}[\hat{\tau}(x) \mid X = x, W = 0] 
= \mathbb{E}[\hat{\tau}(x) \mid X = x, W = 1].
\]
Any deviation from this equality suggests that the estimator is implicitly 
learning different functions depending on the group assignment, which 
contradicts the core assumption of a well-defined, assignment-invariant 
treatment effect.

\paragraph{Discrepancy Ratio.}
To quantify this inconsistency, we introduce a metric called the 
\emph{discrepancy ratio}, which measures the extent to which CATE estimators 
produce significantly different predictions for the same instance depending 
on its treatment assignment during training.

To compute this metric, we utilize a synthetic dataset \citep{Zhao_2022} 
in which both potential outcomes and the ground-truth \(\tau(x)\) are known. 
We repeatedly train each CATE estimator 30 times on the same dataset, 
but with randomized treatment assignments for all training instances in 
each run. For each instance \(x\), we collect the predicted treatment effects 
\(\hat{\tau}(x)\) separately for when it was assigned to the treatment group 
(\(W = 1\)) versus the control group (\(W = 0\)).

We then conduct an independent t-test for each instance to determine whether 
the predictions under the two assignment groups are significantly different. 
The discrepancy ratio is defined as the proportion of instances where the 
difference is statistically significant (p-value \(< 0.05\)).

This metric provides a direct way to measure how often estimators learn 
assignment-dependent treatment effect functions, thereby violating 
assignment-invariance and introducing additional variance.

\begin{figure*}[ht]
  \centering
  \subfloat[Discrepancy ratio vs. data size]{
    \includegraphics[width=0.48\linewidth]{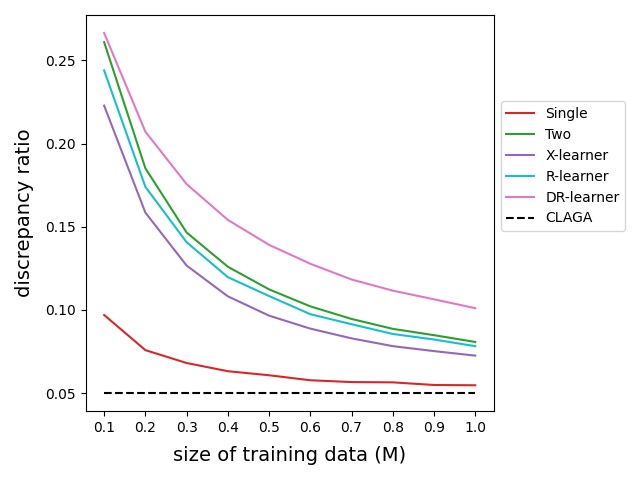}
    \label{fig:data_size}
  }
  \hfill
  \subfloat[Discrepancy ratio vs. model complexity]{
    \includegraphics[width=0.48\linewidth]{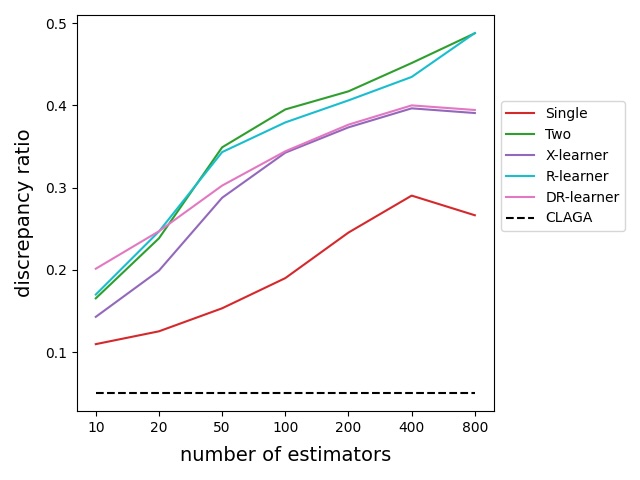}
    \label{fig:model_complexity}
  }
  \caption{(a) The discrepancy ratio decreases as data size increases, 
  suggesting reduced overfitting. 
  (b) The discrepancy ratio increases with higher model complexity, 
  reflecting greater overfitting risk. 
  In both figures, the black dotted line represents our proposed method, CLAGA, 
  which achieves a consistently lower discrepancy ratio.}
  \label{fig:biased_learning}
\end{figure*}

\paragraph{Empirical Observations.}
As shown in Figure~\ref{fig:biased_learning}, 
the discrepancy ratio exhibits two key trends across existing estimators.
In Figure~\ref{fig:data_size}, we observe that the ratio decreases as the 
training data size increases, which aligns with the expectation that more 
data mitigates overfitting to group-specific targets. 
However, even at a scale of 1 million samples, many estimators still exhibit 
substantial inconsistency.

Figure~\ref{fig:model_complexity} further reveals that as model complexity 
increases, the discrepancy ratio rises, indicating that complex models are 
more susceptible to overfitting to assignment-specific noise. 
This reinforces the intuition that over-parameterized models, when trained 
on partial observations, can capture spurious patterns that depend on group 
assignments—despite the fact that the true \(\tau(x)\) should remain invariant.

To address this issue, in Section~\ref{Sec: relabeling process} we will propose a general procedure called 
\emph{Consistent Labeling Across Group Assignments} (CLAGA), which can be 
applied to any CATE estimation method. As indicated by the black dotted line 
in both subfigures, CLAGA consistently reduces the discrepancy ratio across 
data scales and model complexities, demonstrating its ability to mitigate 
assignment-specific overfitting.

\section{IMPACT OF INCONSISTENT LEARNING}
\label{Sec:error-term-breakdown}

\subsection{Error Decomposition}

In this section, we analyze how inconsistent learning across group assignments impacts the accuracy of CATE estimation. Specifically, we focus on the precision in estimating heterogeneous treatment effects, as measured by the Precision in Estimation of Heterogeneous Effect (PEHE) \citep{hill2011bayesian}, which is defined as:
\begin{equation}
  \text{PEHE} := \mathbb{E}_{x} \left[ \mathbb{E}_{M, W}\left[(\hat{\tau}(x) - \tau(x))^2 \mid X = x\right] \right].
\end{equation}

Here, \(M\) denotes the randomness introduced by the model training process (e.g., due to different initializations or optimization stochasticity), and \(W\) refers to the randomness in treatment group assignment. For clarity, we omit the explicit dependence on covariates \(x\) and conduct the following analysis conditioned on a fixed value of \(x\) in the remainder of this section. The extension to the full PEHE over the marginal distribution of \(X\) can be recovered by taking the outer expectation over \(x\).

To understand how different sources contribute to PEHE, we decompose the squared error \(\mathbb{E}_{M,W}[(\hat{\tau} - \tau)^2]\) by expressing the model prediction \(\hat{\tau}\) around the group-dependent learning target \(\Tilde{\tau}\), and subsequently decomposing the surrogate error around the true treatment effect \(\tau\). This yields the following decomposition (full derivation in Appendix):

\begin{equation}
\label{eq:decom_summary}
\begin{aligned}
  \mathbb{E}_{M, W}[(\hat{\tau} - \tau)^2] &= 
  \underbrace{\mathbb{E}_{M, W}[(\hat{\tau} - \Tilde{\tau}^{(W)})^2]}_{\text{Model Error}} \\
  &\quad - \underbrace{2 \mathbb{E}_{M, W}[(\tau - \Tilde{\tau}^{(W)})(\hat{\tau} - \Tilde{\tau}^{(W)})]}_{\text{Model-Target Covariance}} \\
  &\quad + \underbrace{(1 - \pi) \text{Var}[\Tilde{\tau}^{(0)}] + \pi \text{Var}[\Tilde{\tau}^{(1)}]}_{\text{Group Assignment Weighted Variance}} \\
  &\quad + \underbrace{\pi (1 - \pi)(\mathbb{E}[\Tilde{\tau}^{(0)}] - \mathbb{E}[\Tilde{\tau}^{(1)}])^2}_{\text{Inconsistency Across Group Assignments}} \\
  &\quad + \underbrace{\mathbb{E}_W[(\tau - \Tilde{\tau}^{(W)})^2]}_{\text{Bias of Learning Target}},
\end{aligned}
\end{equation}

where

\begin{itemize}
    \item \(\Tilde{\tau}\) refers to the learning target used by the model for a particular instance. It may depend on the observed treatment assignment, \(W\).
    \item \(\Tilde{\tau}^{(0)}\) and \(\Tilde{\tau}^{(1)}\) represent the training targets under the control and treatment assignments, respectively.
    \item \(\pi = \Pr(W = 1)\) is the treatment probability.
\end{itemize}

Each term reflects a distinct source of error:

\begin{itemize}
  \item \textbf{Model Error:} The expected squared error between model prediction \(\hat{\tau}\) and the surrogate label \(\Tilde{\tau}\), capturing the model's inability to fit its assigned target.

  \item \textbf{Model-Target Covariance:} The interaction between surrogate bias \((\tau - \Tilde{\tau})\) and model error \((\hat{\tau} - \Tilde{\tau})\). A negative value suggests the model compensates for the surrogate bias, partially correcting toward the true \(\tau\).

  \item \textbf{Group Assignment Weighted Variance:} The intra-group variance of surrogate targets \(\Tilde{\tau}^{(0)}\) and \(\Tilde{\tau}^{(1)}\), weighted by group assignment probabilities. High within-group variance leads to uncertainty in supervision signals.

  \item \textbf{Inconsistency Across Group Assignments:} The squared difference between the group-conditional mean surrogate labels. This reflects how differently an estimator learns for the same instance under different group assignments.

  \item \textbf{Bias of Learning Target:} The average squared difference between surrogate label \(\Tilde{\tau}\) and the true effect \(\tau\), reflecting structural bias induced by the algorithm’s label construction design.
\end{itemize}

\subsection{Insights into the Error Components}
\label{subsec:component category}

We can categorize the error components in Equation~\ref{eq:decom_summary} into two groups: those that depend on the model training process and those that do not. Specifically, we classify any component involving \(\hat{\tau}\) as training-dependent, and those involving only \(\tilde{\tau}\) or \(\tau\) as independent of model training.

\subsubsection{Error components unrelated to the training process}
\label{subsubsec: error components unrelated to training}

The following components are independent of the model training process and arise solely from the construction of the learning target \(\tilde{\tau}^{(W)}\), conditioned on treatment assignment \(W\):

\begin{itemize}
  \item \(\mathbb{E}_W[(\tau - \Tilde{\tau}^{(W)})^2]\)
  \item \((1 - \pi)\, \text{Var}[\Tilde{\tau}^{(0)}] + \pi\, \text{Var}[\Tilde{\tau}^{(1)}]\)
  \item \(\pi (1 - \pi)\, (\mathbb{E}[\Tilde{\tau}^{(0)}] - \mathbb{E}[\Tilde{\tau}^{(1)}])^2\)
\end{itemize}

For convenience, we define:

\begin{itemize}
  \item \(\text{WVG}(\tilde{\tau})\): weighted variance across group assignment
  \item \(\text{SDMG}(\tilde{\tau})\): squared difference of groupwise means
\end{itemize}

These components depend only on the treatment assignment \(W\) and the algorithm's internal design of \(\tilde{\tau}\), but not on any specific model training process. Consequently, their magnitudes cannot be mitigated by better model training or hyperparameter tuning—they must be addressed through improved learning target construction.

For instance, \(\mathbb{E}_W[(\tau - \tilde{\tau}^{(W)})^2]\) vanishes when the learning target \(\tilde{\tau}\) is unbiased for \(\tau\). Meanwhile, \(\text{WVG}(\tilde{\tau})\) increases when the surrogate targets exhibit higher variance within each group assignment, and \(\text{SDMG}(\tilde{\tau})\) increases when their groupwise means diverge.

\begin{remark}
\(\text{WVG}(\tilde{\tau})\) decreases as the variance of \(\tilde{\tau}^{(0)}\) and \(\tilde{\tau}^{(1)}\) becomes smaller.
\end{remark}

\begin{remark}
\(\text{SDMG}(\tilde{\tau}) = 0\) when \(\mathbb{E}[\tilde{\tau}^{(0)}] = \mathbb{E}[\tilde{\tau}^{(1)}]\), indicating consistent learning targets across group assignments.
\end{remark}

\subsubsection{Error components related to the training process}
\label{subsubsec:error-components-training}

The remaining two components involve \(\hat{\tau}\), the model prediction, and are therefore sensitive to the randomness introduced by the training process \(M\):

\begin{itemize}
  \item \(\mathbb{E}_{M, W}[(\hat{\tau} - \Tilde{\tau}^{(W)})^2]\)
  \item \(-2\, \mathbb{E}_{M, W}[(\tau - \Tilde{\tau}^{(W)})(\hat{\tau} - \Tilde{\tau}^{(W)})]\)
\end{itemize}

The first term measures how well the model fits the surrogate label \(\Tilde{\tau}\), averaged over training and treatment randomness. The second is a covariance term between the surrogate label's bias and the model's prediction error. It reflects whether the model compensates for surrogate bias or exacerbates it.

To understand its contribution, we derive an upper bound using the Cauchy–Schwarz inequality:

\begin{equation}
\begin{aligned}
\left| -2\mathbb{E}_{M, W}[(\tau - \Tilde{\tau}^{(W)})(\hat{\tau} - \Tilde{\tau}^{(W)})] \right| \\
\leq 2\sqrt{\mathbb{E}_{M, W}[(\tau - \Tilde{\tau}^{(W)})^2]} &\cdot \sqrt{\mathbb{E}_{M, W}[(\hat{\tau} - \Tilde{\tau}^{(W)})^2]}.
\end{aligned}
\end{equation}

This shows that the magnitude of this term is upper bounded by the geometric mean of target bias and model error. When both are small—due to well-designed surrogate labels and a well-trained model—this term is also negligible.

\section{CONSISTENT LABELING ACROSS GROUP ASSIGNMENTS}
\label{Sec: relabeling process}

\begin{figure*}[t]
\vskip 0.2in
\begin{center}
\centerline{\includegraphics[width=1\textwidth]{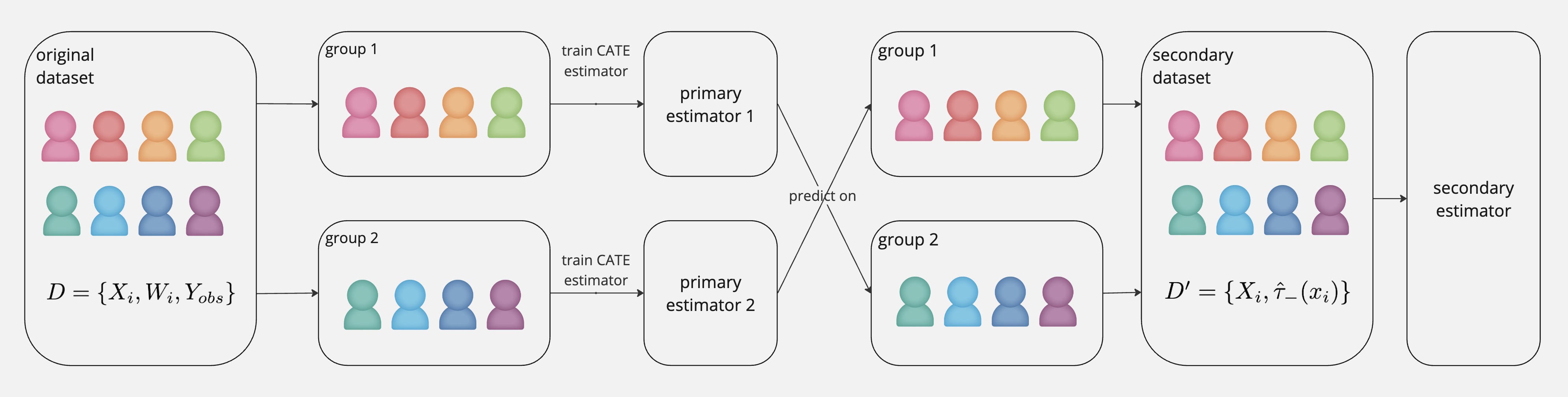}}
\caption{An overview of CLAGA with K=2.}
\label{relabeling process}
\end{center}
\vskip -0.2in
\end{figure*}

\subsection{Motivation}
In Section \ref{subsec:component category}, we identified two types of error components. While model training techniques can help reduce the errors related to the training process, our focus here is on improving the inherent errors caused by the algorithm design, as discussed in Section~\ref{subsubsec: error components unrelated to training}. We propose a general method that can be applied to any CATE estimation algorithm to effectively mitigate these errors.

\subsection{Method}
We introduce a method called \emph{consistent labeling across group assignments} (CLAGA), which aims to generate a less variable learning target, addressing both intra-group and inter-group assignment variance. By reducing intra-group variance, we lower the error term \(\text{WVG}(\Tilde{\tau})\), and by reducing inter-group variance, we lower the error term \(\text{SDMG}(\Tilde{\tau})\).

The process begins by partitioning the training dataset into \(K\) folds. We then train \(K\) primary CATE estimators, each trained on \(K-1\) folds, leaving one fold out. In each primary estimator, the left-out fold is excluded from the training process and is used solely for generating out-of-sample predictions later. After training the primary estimators, for each instance \(X_i\), we derive its out-of-sample prediction using the estimator that did not train on \(X_i\), with the prediction denoted as \(\hat{\tau}_-(X_i)\). Finally, we use the new dataset \(\mathcal{D}' = \{(X_i, \hat{\tau}_-(X_i))\}\) to train a regression model as the secondary CATE estimator. Figure~\ref{relabeling process} provides an overview of CLAGA with \(K=2\). The detailed steps of CLAGA are summarized in Algorithm~\ref{app:claga_algorithm}.

\begin{algorithm}[h]
    \SetAlgoLined
    \KwData{Training dataset \(\mathcal{D} = \{(X_i, W_i, Y_i)\}\), CATE estimation algorithm \(\mathbf{G}\), number of folds \(K\)}
    \KwResult{CATE estimator \(\mathbf{g'}\) using CLAGA}
    
    \caption{Consistent Labeling Across Group Assignments (CLAGA)}
    
    \tcp{Step 1: Partition the dataset}
    Partition \(\mathcal{D}\) into \(K\) folds: \(\mathcal{D}_1, \mathcal{D}_2, \dots, \mathcal{D}_K\)
    
    \tcp{Step 2: Train primary estimators on \(K-1\) folds}
    \For{$i=1$ to $K$}{
        Train primary estimator \(\mathbf{g}_i\) using \(\mathcal{D} - \mathcal{D}_i\) (i.e., all folds except \(\mathcal{D}_i\)) \\
        Generate out-of-sample predictions \(\hat{\tau}_-(X_j)\) for each instance \(X_j \in \mathcal{D}_i\)
    }
    
    \tcp{Step 3: Create relabeled dataset}
    Create new dataset \(\mathcal{D}' = \{(X_i, \hat{\tau}_-(X_i))\}\), where \(\hat{\tau}_-(X_i)\) is the out-of-sample prediction from the primary estimator
    
    \tcp{Step 4: Train secondary estimator}
    Train a regression model \(\mathbf{g'}\) on the entire relabeled dataset \(\mathcal{D}'\)
    
    \Return Final CATE estimator \(\mathbf{g'}\)
\label{app:claga_algorithm}
\end{algorithm}

\subsection{Insight}
CLAGA introduces a new learning target, \(\Tilde{\tau}'\), which is the out-of-sample prediction from the primary CATE estimators, \(\hat{\tau}_-\). Since \(\hat{\tau}_-\) (and thus \(\Tilde{\tau}'\)) does not depend on the group assignment, we have \(\mathbb{E}[\Tilde{\tau}'^{(0)}] = \mathbb{E}[\Tilde{\tau}'^{(1)}]\), meaning that \(\text{SDMG}(\Tilde{\tau}') = 0\). Additionally, since the learning targets for both groups become indistinguishable, \(\text{WVG}(\Tilde{\tau}')\) reduces to \(\text{Var}_W[\Tilde{\tau}']\). To further minimize \(\text{Var}_W[\Tilde{\tau}']\), various machine learning techniques can be applied to make the primary estimators' predictions more consistent. For example, theoretically, if we perform infinite ensemble learning for primary CATE estimators, \(\text{Var}_W[\Tilde{\tau}']\) approaches zero.

\begin{remark}
With CLAGA, \(\text{SDMG}(\Tilde{\tau})\) becomes zero.
\end{remark}
\begin{remark}
With CLAGA, \(\text{WVG}(\Tilde{\tau})\) can be minimized using ensemble predictions from the primary estimators.
\end{remark}

Lastly, we analyze the error term \(\mathbb{E}_W[\tau-\Tilde{\tau}^{(W)}]^2\). Since the primary estimators' predictions are used as secondary labels, indicating \(\mathbb{E}[\Tilde{\tau}'] = \mathbb{E}[\hat{\tau}]\), then if these primary estimators are unbiased (i.e. $\mathbb{E}[\hat{\tau}] = \mathbb{E}[\Tilde{\tau}]$), it follows that \(\mathbb{E}[\Tilde{\tau}'] = \mathbb{E}[\hat{\tau}] = \mathbb{E}[\Tilde{\tau}]\). Therefore, the error term \(\mathbb{E}_W[\tau-\Tilde{\tau}^{(W)}]^2\) remains unchanged, meaning CLAGA does not increase or decrease this particular error component.

\begin{remark}
With CLAGA, the error term \(\mathbb{E}_W[\tau-\Tilde{\tau}^{(W)}]^2\) remains unchanged if unbiased primary estimators are used.
\end{remark}

In summary, our proposed method effectively reduces the error components unrelated to the model training process. When combined with proper model training, CLAGA benefits the performance of CATE estimation.

\section{EXPERIMENT}
\label{sec:experiment}

In this section, we evaluate the effectiveness of CLAGA. Section~\ref{subsection: exp synthetic} presents the evaluation of multiple algorithms with and without CLAGA on synthetic datasets. In Section~\ref{subsection: exp real-world}, we further assess the benefit of CLAGA on real-world datasets.

\subsection{CLAGA on Synthetic Datasets}
\label{subsection: exp synthetic}

\paragraph{Dataset}
We evaluate the PEHE of various algorithms, both with and without CLAGA, on two public synthetic datasets: ACIC-2016 \citep{dorie2019automated} and ACIC-2018 \citep{shimoni2018benchmarking}. The ACIC-2016 dataset contains 4802 instances with both potential outcomes simulated under 77 different conditions, while ACIC-2018 consists of 24 datasets with sizes ranging between 1000 and 50,000. Additionally, we conduct experiments on the Zenodo dataset \citep{Zhao_2022}, as discussed in Section~\ref{subsec:biased learning}, to examine the effect of CLAGA with respect to data size and model complexity.

\paragraph{Experimental Setup}
We use five CATE estimation algorithms, as discussed in Section~\ref{Sec: related work: Algorithms for CATE estimation}, including the Single-model approach \citep{lo2002true}, the Two-model approach \citep{hansotia2002incremental, radcliffe2007using}, X-learner \citep{kunzel2019metalearners}, R-learner \citep{nie2021quasi}, and DR-learner \citep{kennedy2020optimal}. For each algorithm, we train models both with and without CLAGA and evaluate their PEHE on test data. The algorithms are implemented using the Python package CausalML \citep{chen2020causalml}, with LGBMClassifier and LGBMRegressor \citep{ke2017lightgbm} for all estimators.

For ACIC-2016 and ACIC-2018, due to the relatively small data sizes, we use `n\_estimators=100' and `num\_leaves=32'. For the larger Zenodo dataset, we use `n\_estimators=400' and `num\_leaves=64'. A subsample rate of 0.5 and `subsample\_freq=3' are applied across all datasets, with other hyperparameters left as default. In CLAGA, we use \(K=10\) for ACIC-2016 and ACIC-2018, and \(K=2\) for Zenodo. Each experiment setting is repeated 10 times with different random seeds.

\paragraph{Results}

\begin{table}[t]
\caption{PEHE ratios for each algorithm after applying CLAGA.}
\centering
\small
\begin{tabular}{ccc}
    \toprule
    \multicolumn{1}{l}{Algorithm} &
    \multicolumn{1}{c}{ACIC-2016}    &
    \multicolumn{1}{c}{ACIC-2018}    \\ 
    \midrule
    Single-model  & 0.9797\begin{tiny}$\pm$0.0610\end{tiny} & 0.6036\begin{tiny}$\pm$0.3043\end{tiny} \\
    Two-model     & 0.8907\begin{tiny}$\pm$0.0941\end{tiny} & 0.5009\begin{tiny}$\pm$0.2862\end{tiny} \\
    X-learner     & 1.0034\begin{tiny}$\pm$0.0692\end{tiny} & 0.6277\begin{tiny}$\pm$0.2550\end{tiny} \\
    R-learner     & 0.8023\begin{tiny}$\pm$0.0973\end{tiny} & 0.4622\begin{tiny}$\pm$0.2575\end{tiny} \\
    DR-learner    & 0.9340\begin{tiny}$\pm$0.0822\end{tiny} & 0.4919\begin{tiny}$\pm$0.2586\end{tiny} \\
    \midrule
    Avg. PEHE Reduction & 7.8\%  & 46.3\% \\
    \bottomrule
\end{tabular}
\label{table:performance-result-synthetic}
\end{table}

\begin{figure*}[t]
  \centering
  \subfloat[PEHE vs. Data Size]{\includegraphics[width=0.48\textwidth]{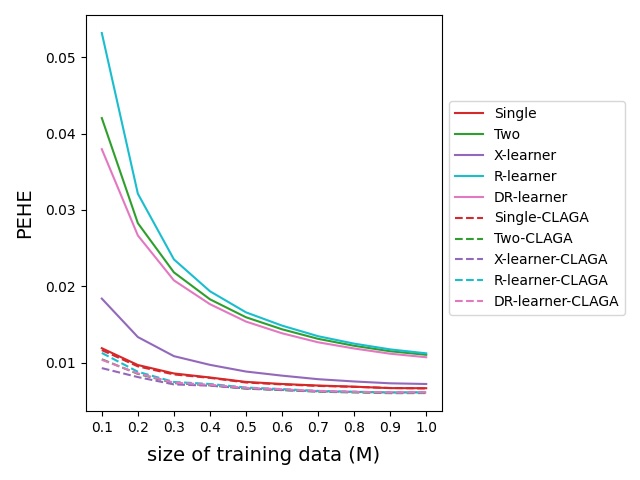}\label{fig:pehe_data}}
  \hfill
  \subfloat[PEHE vs. Model Complexity]{\includegraphics[width=0.48\textwidth]{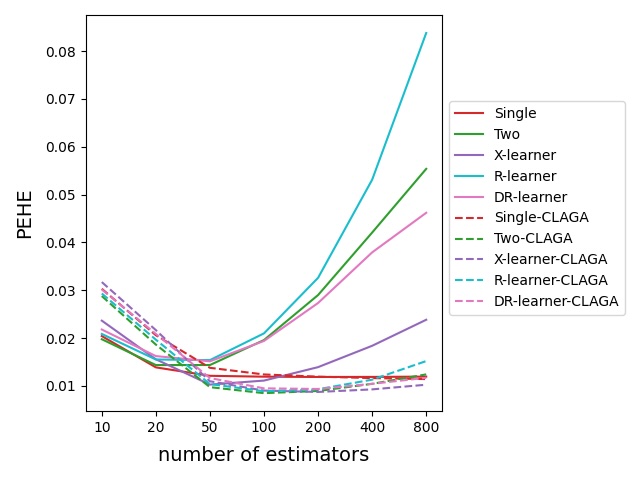}\label{fig:pehe_complexity}}
  \caption{PEHE results across data sizes and model complexities with and without CLAGA.}
  \label{fig:zenodo_synthetic}
\end{figure*}

Table~\ref{table:performance-result-synthetic} shows the PEHE ratio for each algorithm after applying CLAGA, relative to the original setting (without CLAGA). A ratio of less than 1 indicates improved performance, meaning the application of CLAGA reduced PEHE. The results are averaged across all simulation conditions for each dataset.
We observe that CLAGA yields substantial improvements in most settings for both ACIC-2016 and ACIC-2018. Figure~\ref{fig:zenodo_synthetic} shows the impact of CLAGA on PEHE across different data sizes and model complexities on the Zenodo dataset. In Figure~\ref{fig:zenodo_synthetic}(a), CLAGA consistently improves performance across varying data sizes, as indicated by lower PEHE values compared to the baseline. In In Figure~\ref{fig:zenodo_synthetic}(b), we observe that CLAGA is especially beneficial when the model complexity increases, where models are more prone to overfitting. However, we also find that CLAGA only begins to provide benefits beyond a certain level of model complexity. We argue that for models that are too weak, the instability of \(\Tilde{\tau}'\) can lead to negative effects when applying CLAGA. In summary, CLAGA reduces PEHE when applied with appropriate model training settings, highlighting its effectiveness, especially in scenarios with limited data or higher model complexity.

\subsection{CLAGA on Real-world Datasets}
\label{subsection: exp real-world}

\paragraph{Datasets}
We evaluate CLAGA on three real-world datasets: x5 \citep{x5}, Lenta \citep{lenta}, and Criteo \citep{Diemert2018}. In each dataset, the treatment corresponds to either an advertisement or a communication sent to a customer, and the response is measured as either a user visit or a purchase. The Criteo dataset includes two types of responses, so we treat it as two separate versions: Criteo-cv and Criteo-visit, where the response is either user conversion or user visit, respectively.

\paragraph{Experimental Setup}
The general setup follows that of Section~\ref{subsection: exp synthetic}. For x5 and Lenta, we use `n\_estimators=100' and `num\_leaves=64', while for Criteo datasets, we use `n\_estimators=500' and `num\_leaves=256', reflecting the larger dataset size. For CLAGA, we set \(K=2\) for all datasets.

\paragraph{Evaluation Metric}
Since real-world datasets lack ground truth for both potential outcomes, we use the area under the uplift curve (AUUC) \citep{rzepakowski2010decision} as our evaluation metric, which measures the ranking performance of treatment effects across instances. AUUC integrates the uplift curve, which plots cumulative uplift against the cumulative population targeted by an intervention. In practice, AUUC is often preferred in real-world applications, as practitioners are typically more interested in the relative ranking of treatment effects than the absolute treatment effect values. A higher AUUC indicates that the model successfully ranks instances with higher treatment effects.

\paragraph{Results}
\begin{table*}[t!]
\caption{AUUC results for each algorithm, with and without CLAGA, on four real-world datasets. "Avg. Imp." represents the average improvement of applying CLAGA in each dataset.}
\scriptsize
\resizebox{\textwidth}{!}{ 
\begin{tabular}{lcccccccc}
    \toprule
    Algorithm & \multicolumn{2}{c}{x5} & \multicolumn{2}{c}{Lenta} & \multicolumn{2}{c}{Criteo-visit} & \multicolumn{2}{c}{Criteo-cv} \\ 
    \cmidrule(lr){2-3} \cmidrule(lr){4-5} \cmidrule(lr){6-7} \cmidrule(lr){8-9}
    & Vanilla & CLAGA & Vanilla & CLAGA & Vanilla & CLAGA & Vanilla & CLAGA \\
    \midrule
    Single & 0.0505\begin{tiny}$\pm$0.0015\end{tiny} & \textbf{0.0540}\begin{tiny}$\pm$0.0011\end{tiny} & \textbf{0.0135}\begin{tiny}$\pm$0.0007\end{tiny} & 0.0132\begin{tiny}$\pm$0.0007\end{tiny} & 0.0302\begin{tiny}$\pm$0.0001\end{tiny} & \textbf{*0.0309}\begin{tiny}$\pm$0.0001\end{tiny} & 0.00308\begin{tiny}$\pm$0.00004\end{tiny} & \textbf{0.00322}\begin{tiny}$\pm$0.00009\end{tiny} \\
    Two & 0.0537\begin{tiny}$\pm$0.0011\end{tiny} & \textbf{0.0542}\begin{tiny}$\pm$0.0008\end{tiny} & 0.0104\begin{tiny}$\pm$0.0006\end{tiny} & \textbf{0.0128}\begin{tiny}$\pm$0.0005\end{tiny} & 0.0245\begin{tiny}$\pm$0.0001\end{tiny} & \textbf{0.0291}\begin{tiny}$\pm$0.0000\end{tiny} & 0.00321\begin{tiny}$\pm$0.00007\end{tiny} & \textbf{*0.00361}\begin{tiny}$\pm$0.00005\end{tiny} \\
    X-learner & \textbf{0.0539}\begin{tiny}$\pm$0.0005\end{tiny} & 0.0521\begin{tiny}$\pm$0.0010\end{tiny} & 0.0120\begin{tiny}$\pm$0.0005\end{tiny} & \textbf{*0.0150}\begin{tiny}$\pm$0.0006\end{tiny} & 0.0250\begin{tiny}$\pm$0.0001\end{tiny} & \textbf{0.0284}\begin{tiny}$\pm$0.0001\end{tiny} & 0.00295\begin{tiny}$\pm$0.00009\end{tiny} & \textbf{0.00331}\begin{tiny}$\pm$0.00006\end{tiny} \\
    R-learner & 0.0428\begin{tiny}$\pm$0.0012\end{tiny} & \textbf{0.0504}\begin{tiny}$\pm$0.0010\end{tiny} & 0.0115\begin{tiny}$\pm$0.0006\end{tiny} & \textbf{0.0118}\begin{tiny}$\pm$0.0007\end{tiny} & 0.0236\begin{tiny}$\pm$0.0002\end{tiny} & \textbf{0.0277}\begin{tiny}$\pm$0.0001\end{tiny} & 0.00287\begin{tiny}$\pm$0.00009\end{tiny} & \textbf{0.00294}\begin{tiny}$\pm$0.00008\end{tiny} \\
    DR-learner & 0.0388\begin{tiny}$\pm$0.0012\end{tiny} & \textbf{0.0410}\begin{tiny}$\pm$0.0010\end{tiny} & 0.0121\begin{tiny}$\pm$0.0004\end{tiny} & \textbf{0.0122}\begin{tiny}$\pm$0.0005\end{tiny} & 0.0234\begin{tiny}$\pm$0.0002\end{tiny} & \textbf{0.0267}\begin{tiny}$\pm$0.0001\end{tiny} & 0.00267\begin{tiny}$\pm$0.00008\end{tiny} & \textbf{0.00303}\begin{tiny}$\pm$0.00005\end{tiny} \\
    \midrule
    Avg. Imp.& \multicolumn{2}{r}{+5.6\%} & \multicolumn{2}{r}{+9.9\%} & \multicolumn{2}{r}{+13.2\%} & \multicolumn{2}{r}{+9.0\%} \\
    \bottomrule
\end{tabular}
}
\label{table:performance-result-real}
\end{table*}

Table~\ref{table:performance-result-real} presents the AUUC for each algorithm, both with and without CLAGA, on the four real-world datasets. The average improvement in AUUC across all algorithms for each dataset is reported at the bottom. In each pairwise comparison, the algorithm with the better AUUC is highlighted in bold, and the best AUUC for each dataset is marked with an asterisk. Overall, CLAGA demonstrates significant positive improvements in most cases. The best performance is observed when CLAGA is applied across all datasets.

\section{Limitations}
\label{sec:limitations}

While CLAGA has shown promising results in reducing variance-related errors and improving CATE estimation, there are several limitations to our approach that merit consideration.

\paragraph{Dependence on the base CATE estimators}
The effectiveness of CLAGA is closely tied to the primary CATE estimators. Although CLAGA reduces variance-related errors, it relies on the quality of the initial estimators. If these estimators exhibit significant bias or are overly simplified, the overall improvement may be limited. Thus, in scenarios where the primary estimators are weak, CLAGA may not fully realize its potential to enhance CATE estimation.

\paragraph{Increased computational complexity}
The process of training multiple base estimators during the K-fold procedure can significantly increase the overall training time and computational resources, particularly for large datasets or highly complex models. This added computational complexity may limit the practicality of CLAGA in resource-constrained environments where efficiency is a concern.

\paragraph{Theoretical analysis to PEHE}
Our theoretical analysis and error decomposition are specifically focused on minimizing the PEHE as the evaluation metric. While PEHE is widely used in CATE estimation, the error decomposition framework we propose may not directly be guaranteed to extend to other metrics, such as AUUC.

\section{Conclusion}
\label{sec:conclusion}

In this paper, we introduced CLAGA, a novel approach aimed at reducing variance-related errors in CATE estimation by ensuring consistent labeling across group assignments. Through comprehensive experiments on both synthetic and real-world datasets, we demonstrated that CLAGA enhances the performance of various CATE estimation algorithms, particularly in settings where models are susceptible to overfitting.

Our error decomposition framework offers new insights into the key sources of error in CATE estimation, showing how reducing variance can lead to more accurate and reliable treatment effect estimates. By addressing these variance-related challenges, CLAGA  offers fresh perspectives on enhancing model performance in this critical area.

\bibliography{aistats2025}

\begin{thebibliography}{33}
\providecommand{\natexlab}[1]{#1}
\providecommand{\url}[1]{\texttt{#1}}
\expandafter\ifx\csname urlstyle\endcsname\relax
  \providecommand{\doi}[1]{doi: #1}\else
  \providecommand{\doi}{doi: \begingroup \urlstyle{rm}\Url}\fi

\bibitem[Alaa and Van Der~Schaar(2019)]{alaa2019validating}
Ahmed Alaa and Mihaela Van Der~Schaar.
\newblock Validating causal inference models via influence functions.
\newblock In \emph{International Conference on Machine Learning}, pages 191--201. PMLR, 2019.

\bibitem[Bica et~al.(2021)Bica, Alaa, Lambert, and Van Der~Schaar]{bica2021real}
Ioana Bica, Ahmed~M Alaa, Craig Lambert, and Mihaela Van Der~Schaar.
\newblock From real-world patient data to individualized treatment effects using machine learning: current and future methods to address underlying challenges.
\newblock \emph{Clinical Pharmacology \& Therapeutics}, 109\penalty0 (1):\penalty0 87--100, 2021.

\bibitem[Chen et~al.(2020)Chen, Harinen, Lee, Yung, and Zhao]{chen2020causalml}
Huigang Chen, Totte Harinen, Jeong-Yoon Lee, Mike Yung, and Zhenyu Zhao.
\newblock Causalml: Python package for causal machine learning, 2020.

\bibitem[Curth and van~der Schaar(2021)]{curth2021nonparametric}
Alicia Curth and Mihaela van~der Schaar.
\newblock Nonparametric estimation of heterogeneous treatment effects: From theory to learning algorithms.
\newblock In \emph{International Conference on Artificial Intelligence and Statistics}, pages 1810--1818. PMLR, 2021.

\bibitem[{Diemert Eustache, Betlei Artem} et~al.(2018){Diemert Eustache, Betlei Artem}, Renaudin, and Massih-Reza]{Diemert2018}
{Diemert Eustache, Betlei Artem}, Christophe Renaudin, and Amini Massih-Reza.
\newblock A large scale benchmark for uplift modeling.
\newblock In \emph{Proceedings of the AdKDD and TargetAd Workshop, KDD, London,United Kingdom, August, 20, 2018}. ACM, 2018.

\bibitem[Dorie et~al.(2019)Dorie, Hill, Shalit, Scott, and Cervone]{dorie2019automated}
Vincent Dorie, Jennifer Hill, Uri Shalit, Marc Scott, and Dan Cervone.
\newblock Automated versus do-it-yourself methods for causal inference: Lessons learned from a data analysis competition.
\newblock 2019.

\bibitem[Foster et~al.(2011)Foster, Taylor, and Ruberg]{foster2011subgroup}
Jared~C Foster, Jeremy~MG Taylor, and Stephen~J Ruberg.
\newblock Subgroup identification from randomized clinical trial data.
\newblock \emph{Statistics in medicine}, 30\penalty0 (24):\penalty0 2867--2880, 2011.

\bibitem[Gubela et~al.(2019)Gubela, Bequ{\'e}, Lessmann, and Gebert]{gubela2019conversion}
Robin Gubela, Artem Bequ{\'e}, Stefan Lessmann, and Fabian Gebert.
\newblock Conversion uplift in e-commerce: A systematic benchmark of modeling strategies.
\newblock \emph{International Journal of Information Technology \& Decision Making}, 18\penalty0 (03):\penalty0 747--791, 2019.

\bibitem[Gutierrez and G{\'e}rardy(2017)]{gutierrez2017causal}
Pierre Gutierrez and Jean-Yves G{\'e}rardy.
\newblock Causal inference and uplift modelling: A review of the literature.
\newblock In \emph{International Conference on Predictive Applications and APIs}, pages 1--13. PMLR, 2017.

\bibitem[Hansotia and Rukstales(2002)]{hansotia2002incremental}
Behram Hansotia and Brad Rukstales.
\newblock Incremental value modeling.
\newblock \emph{Journal of Interactive Marketing}, 16\penalty0 (3):\penalty0 35, 2002.

\bibitem[Hill(2011)]{hill2011bayesian}
Jennifer~L Hill.
\newblock Bayesian nonparametric modeling for causal inference.
\newblock \emph{Journal of Computational and Graphical Statistics}, 20\penalty0 (1):\penalty0 217--240, 2011.

\bibitem[Jaskowski and Jaroszewicz(2012)]{jaskowski2012uplift}
Maciej Jaskowski and Szymon Jaroszewicz.
\newblock Uplift modeling for clinical trial data.
\newblock In \emph{ICML Workshop on Clinical Data Analysis}, volume~46, 2012.

\bibitem[Johansson et~al.(2016)Johansson, Shalit, and Sontag]{johansson2016learning}
Fredrik Johansson, Uri Shalit, and David Sontag.
\newblock Learning representations for counterfactual inference.
\newblock In \emph{International conference on machine learning}, pages 3020--3029. PMLR, 2016.

\bibitem[Ke et~al.(2017)Ke, Meng, Finley, Wang, Chen, Ma, Ye, and Liu]{ke2017lightgbm}
Guolin Ke, Qi~Meng, Thomas Finley, Taifeng Wang, Wei Chen, Weidong Ma, Qiwei Ye, and Tie-Yan Liu.
\newblock Lightgbm: A highly efficient gradient boosting decision tree.
\newblock \emph{Advances in neural information processing systems}, 30:\penalty0 3146--3154, 2017.

\bibitem[Kennedy(2020)]{kennedy2020optimal}
Edward~H Kennedy.
\newblock Optimal doubly robust estimation of heterogeneous causal effects.
\newblock \emph{arXiv preprint arXiv:2004.14497}, 2020.

\bibitem[Knaus(2022)]{knaus2022double}
Michael~C Knaus.
\newblock Double machine learning-based programme evaluation under unconfoundedness.
\newblock \emph{The Econometrics Journal}, 25\penalty0 (3):\penalty0 602--627, 2022.

\bibitem[K{\"u}nzel et~al.(2019)K{\"u}nzel, Sekhon, Bickel, and Yu]{kunzel2019metalearners}
S{\"o}ren~R K{\"u}nzel, Jasjeet~S Sekhon, Peter~J Bickel, and Bin Yu.
\newblock Metalearners for estimating heterogeneous treatment effects using machine learning.
\newblock \emph{Proceedings of the national academy of sciences}, 116\penalty0 (10):\penalty0 4156--4165, 2019.

\bibitem[Lenta(2020)]{lenta}
Microsoft Lenta.
\newblock Bigtarget hackathon hosted by lenta and microsoft.
\newblock \url{https://www.kaggle.com/datasets/mrmorj/bigtarget}, 2020.

\bibitem[Lo(2002)]{lo2002true}
Victor~SY Lo.
\newblock The true lift model: a novel data mining approach to response modeling in database marketing.
\newblock \emph{ACM SIGKDD Explorations Newsletter}, 4\penalty0 (2):\penalty0 78--86, 2002.

\bibitem[Nie and Wager(2021)]{nie2021quasi}
Xinkun Nie and Stefan Wager.
\newblock Quasi-oracle estimation of heterogeneous treatment effects.
\newblock \emph{Biometrika}, 108\penalty0 (2):\penalty0 299--319, 2021.

\bibitem[Radcliffe(2007)]{radcliffe2007using}
Nicholas Radcliffe.
\newblock Using control groups to target on predicted lift: Building and assessing uplift model.
\newblock \emph{Direct Marketing Analytics Journal}, pages 14--21, 2007.

\bibitem[Rubin(1974)]{rubin1974estimating}
Donald~B Rubin.
\newblock Estimating causal effects of treatments in randomized and nonrandomized studies.
\newblock \emph{Journal of educational Psychology}, 66\penalty0 (5):\penalty0 688, 1974.

\bibitem[Rubin(2005)]{rubin2005causal}
Donald~B Rubin.
\newblock Causal inference using potential outcomes: Design, modeling, decisions.
\newblock \emph{Journal of the American Statistical Association}, 100\penalty0 (469):\penalty0 322--331, 2005.

\bibitem[Rzepakowski and Jaroszewicz(2010)]{rzepakowski2010decision}
Piotr Rzepakowski and Szymon Jaroszewicz.
\newblock Decision trees for uplift modeling.
\newblock In \emph{2010 IEEE International Conference on Data Mining}, pages 441--450. IEEE, 2010.

\bibitem[Schuler et~al.(2017)Schuler, Jung, Tibshirani, Hastie, and Shah]{schuler2017synth}
Alejandro Schuler, Ken Jung, Robert Tibshirani, Trevor Hastie, and Nigam Shah.
\newblock Synth-validation: Selecting the best causal inference method for a given dataset.
\newblock \emph{arXiv preprint arXiv:1711.00083}, 2017.

\bibitem[Shimoni et~al.(2018)Shimoni, Yanover, Karavani, and Goldschmnidt]{shimoni2018benchmarking}
Yishai Shimoni, Chen Yanover, Ehud Karavani, and Yaara Goldschmnidt.
\newblock Benchmarking framework for performance-evaluation of causal inference analysis.
\newblock \emph{arXiv preprint arXiv:1802.05046}, 2018.

\bibitem[Wendling et~al.(2018)Wendling, Jung, Callahan, Schuler, Shah, and Gallego]{wendling2018comparing}
Thierry Wendling, Kenneth Jung, Alison Callahan, Alejandro Schuler, Nigam~H Shah, and Blanco Gallego.
\newblock Comparing methods for estimation of heterogeneous treatment effects using observational data from health care databases.
\newblock \emph{Statistics in medicine}, 37\penalty0 (23):\penalty0 3309--3324, 2018.

\bibitem[X5-Retail-Group(2019)]{x5}
X5-Retail-Group.
\newblock Data of x5 retailhero uplift modeling competition.
\newblock \url{https://ods.ai/competitions/x5-retailhero-uplift-modeling/data}, 2019.

\bibitem[Xie et~al.(2012)Xie, Brand, and Jann]{sociology_ref}
Yu~Xie, Jennie~E Brand, and Ben Jann.
\newblock Estimating heterogeneous treatment effects with observa- tional data.
\newblock \emph{Sociological methodology}, 42\penalty0 (1):\penalty0 314--347, 2012.

\bibitem[Yao et~al.(2018)Yao, Li, Li, Huai, Gao, and Zhang]{yao2018representation}
Liuyi Yao, Sheng Li, Yaliang Li, Mengdi Huai, Jing Gao, and Aidong Zhang.
\newblock Representation learning for treatment effect estimation from observational data.
\newblock \emph{Advances in neural information processing systems}, 31, 2018.

\bibitem[Zhang et~al.(2021{\natexlab{a}})Zhang, Li, and Liu]{zhang2021unified}
Weijia Zhang, Jiuyong Li, and Lin Liu.
\newblock A unified survey of treatment effect heterogeneity modelling and uplift modelling.
\newblock \emph{ACM Computing Surveys (CSUR)}, 54\penalty0 (8):\penalty0 1--36, 2021{\natexlab{a}}.

\bibitem[Zhang et~al.(2021{\natexlab{b}})Zhang, Liu, and Li]{zhang2021treatment}
Weijia Zhang, Lin Liu, and Jiuyong Li.
\newblock Treatment effect estimation with disentangled latent factors.
\newblock In \emph{Proceedings of the AAAI Conference on Artificial Intelligence}, volume~35, pages 10923--10930, 2021{\natexlab{b}}.

\bibitem[Zhao(2022)]{Zhao_2022}
Zhenyu Zhao.
\newblock Synthetic data for uplift modeling and heterogenous treatment effect with known counterfactuals and ite, March 2022.
\newblock URL \url{https://zenodo.org/record/6342552}.

\end{thebibliography}

\clearpage 
\appendix
\newpage
\section{Error Decomposition Derivation}
\label{app:error-decomposition}

In this appendix, we provide the full derivation of the error decomposition discussed in Section~\ref{Sec:error-term-breakdown}. We begin with the precision in estimating heterogeneous effects (PEHE), defined as:
\begin{equation}
  \mathbb{E}_{M, W}[(\hat{\tau}(x) - \tau(x))^2].
\end{equation}

Here, \(\hat{\tau}(x)\) is the model prediction, \(\tau(x)\) is the ground truth treatment effect, \(M\) denotes randomness from model training (e.g., initialization or optimization), and \(W\) is the treatment assignment. For notational simplicity, we drop the dependency on \(x\) and write \(\hat{\tau}\), \(\tau\), and \(\Tilde{\tau}\) instead of \(\hat{\tau}(x)\), \(\tau(x)\), and \(\Tilde{\tau}(x)\), respectively.

Recall that \(\Tilde{\tau}^{(w)}\) denotes the group-specific surrogate label assigned to an instance under treatment assignment \(w \in \{0,1\}\), and \(\Tilde{\tau}\) is the random variable over \(W\) such that:
\[
\Tilde{\tau} = 
\begin{cases}
\Tilde{\tau}^{(0)}, & \text{if } W = 0 \\
\Tilde{\tau}^{(1)}, & \text{if } W = 1.
\end{cases}
\]

\subsection*{Step 1: Decomposition of PEHE}

We begin by introducing an intermediate surrogate \(\Tilde{\tau}\), and expanding the squared error:
\begin{equation}
\label{eq:decom1}
\begin{aligned}
\mathbb{E}_{M, W}[(\hat{\tau} - \tau)^2] 
&= \mathbb{E}_{M, W}[((\hat{\tau} - \Tilde{\tau}) + (\Tilde{\tau} - \tau))^2] \\
&= \mathbb{E}_{M, W}[(\hat{\tau} - \Tilde{\tau})^2] 
    + \mathbb{E}_{M, W}[(\tau - \Tilde{\tau})^2] \\
&\quad + 2 \mathbb{E}_{M, W}[(\hat{\tau} - \Tilde{\tau})(\Tilde{\tau} - \tau)].
\end{aligned}
\end{equation}

Rearranging the terms gives:
\[
\mathbb{E}_{M, W}[(\hat{\tau} - \tau)^2] 
= \mathbb{E}_{M, W}[(\hat{\tau} - \Tilde{\tau})^2]
- 2 \mathbb{E}_{M, W}[(\hat{\tau} - \Tilde{\tau})(\tau - \Tilde{\tau})]
+ \mathbb{E}_{M, W}[(\tau - \Tilde{\tau})^2].
\]

Note that the last term depends only on \(W\) and not on model randomness \(M\), and can be rewritten as:
\[
\mathbb{E}_W[(\tau - \Tilde{\tau})^2].
\]

\subsection*{Step 2: Decomposition of \(\mathbb{E}_W[(\tau - \Tilde{\tau})^2]\)}

We expand this term using the standard identity:
\begin{equation}
\label{eq:decom2}
\begin{aligned}
\mathbb{E}_W[(\tau - \Tilde{\tau})^2]
&= \text{Var}_W[\Tilde{\tau}] + \mathbb{E}_W[\tau - \Tilde{\tau}]^2.
\end{aligned}
\end{equation}

Since \(\tau\) is a deterministic function of \(x\), \(\text{Var}_W[\tau] = 0\) and \(\text{Cov}_W[\tau, \Tilde{\tau}] = 0\). Thus, no cross-term arises from interaction with \(\tau\).

\subsection*{Step 3: Variance of Surrogate Labels}

As \(\Tilde{\tau}\) is determined by \(W\) via a mixture of \(\Tilde{\tau}^{(0)}\) and \(\Tilde{\tau}^{(1)}\), the variance over \(W\) is given by:
\begin{equation}
\label{eq:decom4}
\begin{aligned}
\text{Var}_W[\Tilde{\tau}]
&= (1 - \pi) \text{Var}[\Tilde{\tau}^{(0)}] + \pi \text{Var}[\Tilde{\tau}^{(1)}] \\
&\quad + \pi(1 - \pi)\left(\mathbb{E}[\Tilde{\tau}^{(0)}] - \mathbb{E}[\Tilde{\tau}^{(1)}]\right)^2,
\end{aligned}
\end{equation}
where \(\pi = \Pr(W=1)\) is the marginal treatment probability.

\subsection*{Step 4: Final Decomposition}

Combining the results from Equations~\ref{eq:decom1} to \ref{eq:decom4}, we arrive at:
\begin{equation}
\label{eq:decom-all}
\begin{aligned}
\mathbb{E}_{M, W}[(\hat{\tau} - \tau)^2]
&= \mathbb{E}_{M, W}[(\hat{\tau} - \Tilde{\tau})^2] \\
&\quad - 2 \mathbb{E}_{M, W}[(\hat{\tau} - \Tilde{\tau})(\tau - \Tilde{\tau})] \\
&\quad + (1 - \pi) \text{Var}[\Tilde{\tau}^{(0)}] + \pi \text{Var}[\Tilde{\tau}^{(1)}] \\
&\quad + \pi (1 - \pi)\left(\mathbb{E}[\Tilde{\tau}^{(0)}] - \mathbb{E}[\Tilde{\tau}^{(1)}]\right)^2 \\
&\quad + \left(\mathbb{E}_W[\tau - \Tilde{\tau}]\right)^2.
\end{aligned}
\end{equation}

This aligns directly with the error terms in Section~\ref{Sec:error-term-breakdown}, showing how inconsistency in surrogate label construction contributes directly to the final estimation error.

\end{document}